\documentclass[10pt,twocolumn,letterpaper]{article}

\usepackage[pagenumbers]{cvpr} 

\usepackage[dvipsnames]{xcolor}

\usepackage{kotex}

\usepackage[accsupp]{axessibility} 
\usepackage{setspace}

\definecolor{cvprblue}{rgb}{0.21,0.49,0.74}
\usepackage[pagebackref,breaklinks,colorlinks,citecolor=cvprblue]{hyperref}

\usepackage{fontawesome}
\def\paperID{3770} 
\def\confName{CVPR}
\def\confYear{2024}

\title{Exploiting Style Latent Flows for Generalizing Deepfake Video Detection}

\author{
Jongwook Choi$^{1,2}$ \quad
Taehoon Kim$^1$ \quad
Yonghyun Jeong$^3$ \quad
Seungryul Baek$^4$ \quad
Jongwon Choi$^{1,2}$\thanks{Corresponding author} \\
{\small $^1$Dept. of Advanced Imaging, Chung-Ang Univ, Korea}
{\small $^2$GS. of AI, Chung-Ang Univ, Korea} \\
{\small $^3$Image Vision, NAVER Cloud, Korea}
{\small $^4$AI Graduate School, UNIST, Korea}\\
{\tt\small\{cjw, kimth\}@vilab.cau.ac.kr,
yonghyun.jeong@navercorp.com,
srbaek@unist.ac.kr,
choijw@cau.ac.kr}
}

\AddToShipoutPicture*{%
     \AtTextUpperLeft{%
         \put(0,30){
           \begin{minipage}{\textwidth}
              \footnotesize
              Preprint version; final version will be available at \url{https://openaccess.thecvf.com}\\
              The IEEE / CVF Computer Vision and Pattern Recognition Conference (CVPR) (2024)\\
              Published by: IEEE \& CVF
           \end{minipage}}%
     }%
}

\begin{document}
\maketitle
\begin{abstract}
This paper presents a new approach for the detection of fake videos, based on the analysis of style latent vectors and their abnormal behavior in temporal changes in the generated videos. We discovered that the generated facial videos suffer from the temporal distinctiveness in the temporal changes of style latent vectors, which are inevitable during the generation of temporally stable videos with various facial expressions and geometric transformations. Our framework utilizes the StyleGRU module, trained by contrastive learning, to represent the dynamic properties of style latent vectors. Additionally, we introduce a style attention module that integrates StyleGRU-generated features with content-based features, enabling the detection of visual and temporal artifacts. We demonstrate our approach across various benchmark scenarios in deepfake detection, showing its superiority in cross-dataset and cross-manipulation scenarios. Through further analysis, we also validate the importance of using temporal changes of style latent vectors to improve the generality of deepfake video detection.
\end{abstract}
\section{Introduction}
\label{sec:intro}

Recent generative algorithms are capable of producing high-quality videos; while this advancement has led to social concerns, as it becomes increasingly challenging to distinguish between generated videos and authentic ones.
Generative models have the potential to expedite industries such as entertainment, gaming, fashion, design, and education.
However, their misuse can have adverse effects on society.
The high quality of the videos intensifies the potential for inappropriate utilization of the technique.
To resolve the issue, researchers are actively engaged in developing the deepfake video detection algorithms~\cite{ff++2019, CDF, DFo}.

Early deepfake detection research addressed spatial artifacts such as unnatural aspects~\cite{GramNet2020} and frequency-level checkerboard of the generative model~\cite{marra2019} in the generation of a single frame.
While spatial artifact-based detection methods have shown reasonable performance for single images, they failed in accounting for temporal artifacts in deepfake videos that consist of multiple frames.
To address the limitation, recent studies~\cite{cnn-gru2019, stil2021} integrate temporal cues such as flickering~\cite{FTCN2021, fine-grained2022} and discontinuity~\cite{lipforensics2021} to enhance the accuracy of deepfake video detection task.

\begin{figure}[t]
\centering
\includegraphics[width=1.0\columnwidth]{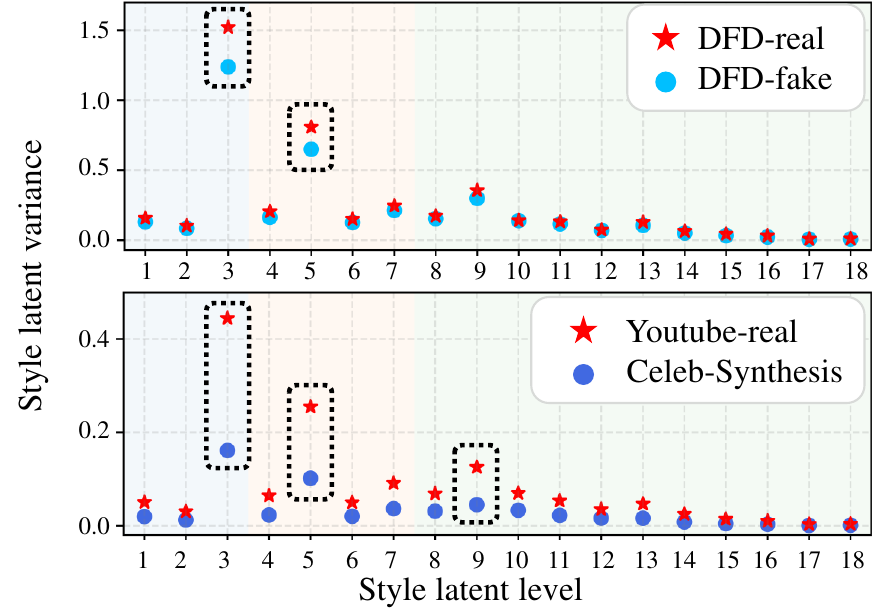} 
\caption{\textbf{Variance of style flow for each style latent level.} The x-axis shows the level of style latent vectors for fine style representations. We noticed that the level-wise differences vary across deepfake domains, but the variance of style latent vectors is particularly lower in certain levels of the style latent vectors for fake videos than in real videos. This happens due to the temporal smoothness of the style latent vectors to create temporally stable deepfake videos, and our results demonstrate that deepfake videos have a distinct variance in style flow compared to real videos.
}
\label{fig1}
\end{figure}

Nevertheless, existing methods~\cite{fwa2018, noisedf2023} that exploit visual and temporal cues of generated videos encountered performance degradation when attempting to identify videos produced by recent deepfake generation algorithms, which effectively suppress visual and temporal artifacts.
Recent observations~\cite{diffusiondetection2022, ojha2023towards} have presented a decline in the effectiveness of visual artifact-based algorithms against newly developed high-quality generation techniques~\cite{dalle2021, diffusion2021}. Temporal artifact-based algorithms~\cite{lipforensics2021} also experienced performance degradation due to the advanced quality of video generation methods~\cite{cnn-gru2019, FTCN2021}.

In this paper, we propose to focus on the suppressed variance in the temporal flow of style latent features which are extracted from generated videos, rather than conventional visual and temporal artifacts.
The style latent feature encodes the facial appearance and facial movements, such as expressions and geometric transformations, to control the synthesis of facial images.
Fig.~\ref{fig1} shows the clear differences in the flow of style latent vectors between the generated and synthetic videos.
We noticed that the suppressed variance of the style latent vector occurs due to the temporal smoothness of facial appearance and movements in the generated videos.
We believe that the flow of style latent vectors has the potential to be a distinctive feature for generalizing deepfake video detection across various generative models.

In order to encode the dynamics of style latent vectors, we introduce the StyleGRU module in our video deepfake detection framework.
This module can effectively capture the temporal variations within style latent vectors, consequently enabling the extraction of robust style-based temporal features.
To increase the representation power of style-based temporal features, we further involve the supervised contrastive learning~\cite{supervised2020} technique in StyleGRU module.
We develop the style attention module which effectively integrates the style-based temporal features with the content features that encode the visual and temporal artifacts.

We perform various experiments to evaluate the proposed algorithm for multiple scenarios such as cross-dataset and cross-manipulation scenarios.
From the experimental results, our method notably surpasses the existing algorithms, achieving a state-of-the-art performance.
Ablation studies are also conducted to show the module-wise effectiveness, and the meaningful analysis experiments confirm that the variations in style latent vector contribute effectively to the robust detection to the various deepfake generation methods.
The contributions of our work are summarized as follows.
\begin{itemize}
\item We propose a novel video deepfake detection framework that is based on the unnatural variation of the style latent vectors.
\item We introduce a StyleGRU module that can encode the variations of style latent vectors with the contrastive learning technique.
\item We propose a style attention module that effectively integrates the features from StyleGRU with the content features capturing conventional visual and temporal artifacts.
\item Our approach demonstrates state-of-the-art performance in various deepfake detection scenarios, including cross-dataset and cross-manipulation settings, which confirms the effectiveness of facial attribute changes for deepfake video detection.
\end{itemize}

\section{Related work}
\begin{figure*}[t]
\centering
\includegraphics[width=1.0\textwidth]{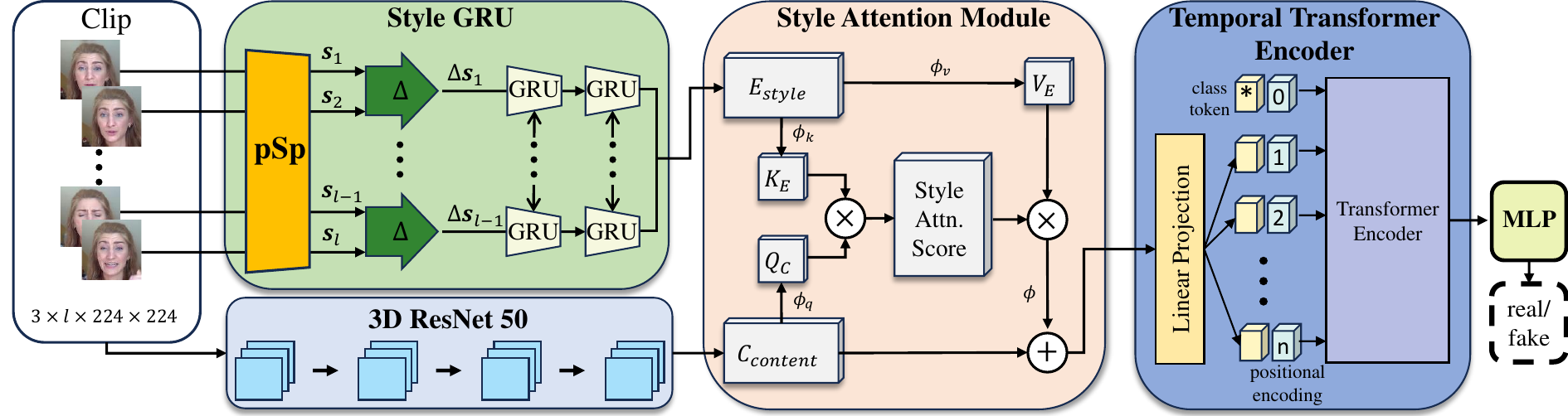} 
\caption{\textbf{Schematic diagram of our entire framework.} For the video deepfake detection, we first extract the style latent vectors from pSp for each individual frame, and then encode their variations using the StyleGRU module into the style-based temporal feature $E_\text{style}$. In parallel, the content feature $C_\text{content}$ is extracted via the 3D ResNet-50 architecture from the video clip. Style Attention Module (SAM) integrates the style-based temporal feature $E_\text{style}$ with the content feature $C_\text{content}$ via the attention mechanism. Finally, Temporal Transformer Encoder (TTE) is applied to map the representation into the binary class of real and fake labels.}
\label{fig2}
\end{figure*}
\noindent\textbf{Deepfake detection.}
In deepfake detection research, there are two main types: image-level and video-level detectors.
The image-level detectors detect fake images by recognizing spatial artifacts of a single frame.
A straightforward approach to detect spatial artifacts involves using Xception model~\cite{ff++2019}, a Convolutional Neural Network (CNN) architecture along with MAT~\cite{9577592} that utilizes an attentional mechanism.
Face X-ray~\cite{facexray2020} proposed a method that uses boundaries between forged faces and backgrounds to capture spatial inconsistency.
Several methods utilized the frequency-level features to emphasize the artifacts from the image generators~\cite{fine-grained2022}.
FrePGAN~\cite{FrePGAN} utilized the frequency map transformed from the original RGB image, and LRL~\cite{chen2021local} integrated RGB images with frequency-aware cues to learn inconsistencies.
Recently, algorithms utilize methods including blending artifacts~\cite{sbi2022, sladd2022, implicit, aunet} or separating elements relevant to detection from irrelevant ones~\cite{implicit, UCF} during their training.
Image-level detectors demonstrate high generalization performance. However, image-based methods have the limitation of not utilizing the temporal cues that inevitably occur during the generation of deepfake videos.

In contrast to image-level detectors, video-level detectors take advantage of temporal information by using multiple frames to detect deepfake videos.
CNN-GRU~\cite{sabir2019recurrent} captured temporal information from the features extracted by CNNs, and LipForensics~\cite{lipforensics2021} utilized the information near the mouth region by employing a network pre-trained on a LipReading dataset. 
Recently, FTCN~\cite{FTCN2021} directly extracted temporal information using 3D CNNs with a spatial kernel size of $1$.
AltFreeze~\cite{altfreezing2023} showcases strong generalization capabilities through its unique approach of independently training spatial and temporal information.
However, the existing methods have focused on low-level features like pixels, so they ignore the temporal changes of high-level contexts such as facial attributes.
To improve detection capabilities, we propose a novel approach that considers temporal features at both low-level and high-level, thereby enhancing the overall context and advancing the state-of-the-art in deepfake detection.

\noindent\textbf{Style Latent Vectors.}
Well-trained Generative Adversarial Network (GAN) models~\cite{StyleGAN1, StyleGAN2, StyleGAN3} successfully generate disentangled latent space. This advancement permits the straightforward manipulation of style latent features to produce desired facial images.
The GAN inversion, initially introduced by \cite{inversion2016}, targets the identification of a latent code within the latent space, which makes it possible to generate an image closely resembling a target one.
Furthermore, various encoders~\cite{zhu2020domain,pSp2021, e4e2021, restyle2021} have been studied to properly utilize the prior knowledge of StyleGAN latent space.
Moreover, GAN inversion methods decoding the style latent features of StyleGAN~\cite{StyleGAN2} are utilized for generating the high-quality face-swap videos.
MegaFS~\cite{MegaFS2021} also inverted both source and target images into the finely-tuned StyleGAN latent space for the purpose of face swap.

Face swapping through GAN inversion offers the advantage of separating facial identity in the latent space and swapping latent features, thereby preserving fidelity in models like StyleGAN~\cite{StyleGAN1}. Furthermore, it is employed for various Deepfake generation techniques, including modifying facial attributes, extending beyond just altering facial identity.
These approaches surpassed the existing method, such as StyleRig~\cite{stylerig2020}, which involved direct modifications to the style latent feature.
The GAN inversion-based face swap methods~\cite{Higres2022, region2022, FGFS2023} continue to demonstrate good performance.
We also employ the style latent features of GAN inversion models to represent the temporal changes of facial attributes. Given that most of the recent GAN inversion models, including those based on the StyleGAN architecture~\cite{StyleGAN1}, utilize style features, a similar situation applies to the construction of deepfake generation models. Hence, we propose the use of StyleGRU, which is well suited for countering attacks on state-of-the-art deepfake detection models.
\section{Methodology}
\begin{figure*}[t]
\centering
\includegraphics[width=1.0\textwidth]{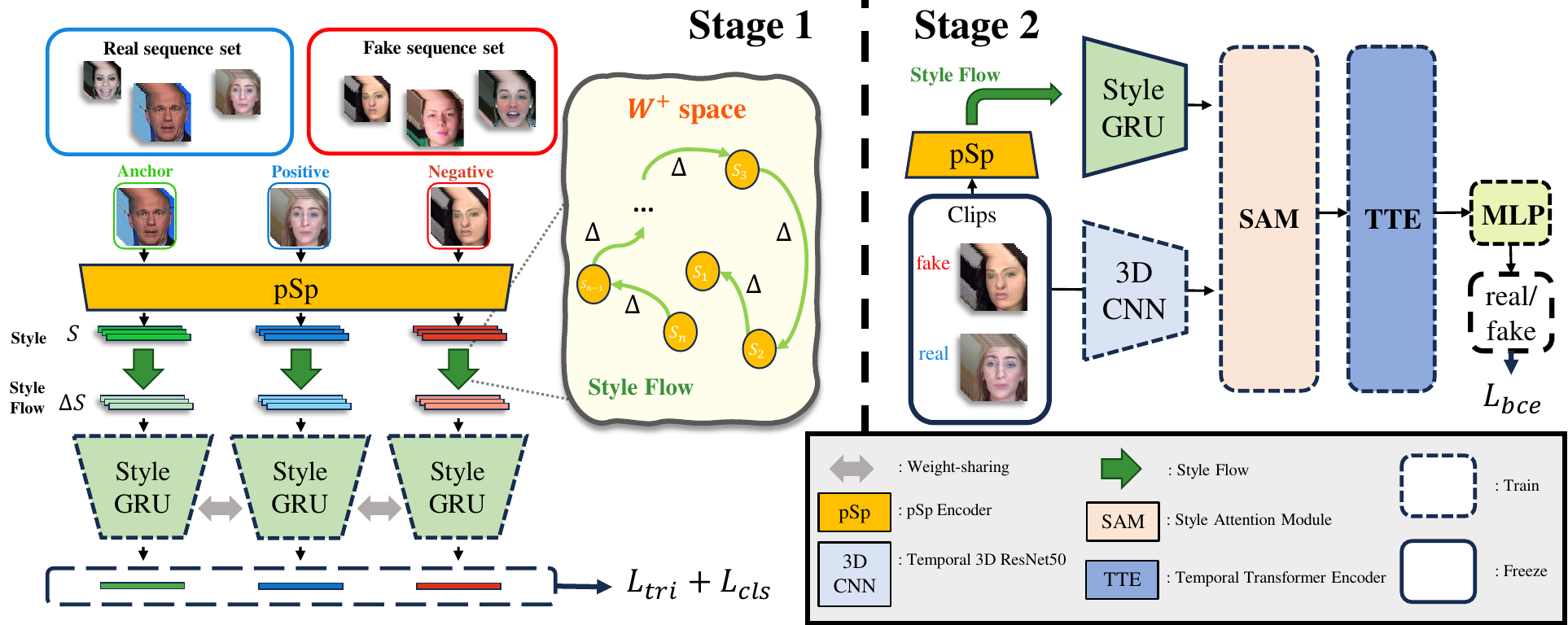} 
\caption{\textbf{Our training procedures.} In stage 1, we train the StyleGRU module using the supervised contrastive learning technique to effectively capture variations of the pSp feature and encode it into the robust style-based temporal feature. In stage 2, we train Style Attention Module (SAM) and Temporal Transformer Encoder (TTE) to integrate the style-based temporal feature and content feature and then map it towards the binary classes (ie. real and fake), respectively.}
\label{fig3}
\end{figure*}

In this section, we propose the video deepfake detection framework that exploits the style latent vectors and captures their characteristics in the generated videos.
Especially, we proposed four distinct modules to achieve the goal: (1) StyleGRU module that extracts the pSp features for each individual frame and encode the temporal variations of the style latent vectors using the GRU into the style-based temporal feature $E_\text{style}$, (2) 3D ResNet-50 module that encodes the video clip into the content feature $C_\text{content}$, (3) Style Attention Module (SAM) that combines the style-based temporal feature $E_\text{style}$ with the content feature $C_\text{content}$ via the attention mechanism, (4) Temporal Transformer Encoder (TTE) module that effectively maps the combined features into the binary classes (i.e. real and fake).
Fig.~\ref{fig2} illustrates the overall flow of our framework.

\subsection{Overall Architecture}

\subsubsection{Input configuration}
We obtain the video clip $I$ that contains a sequence of $l$ consecutive frames extracted from a single video.
Then, we build a clip $I_{p}$ composed of all cropped faces within the input clip $I$ that have been preprocessed with FFHQ~\cite{StyleGAN1} conditions.
The preprocessed video clip $I_p$ is used as an input to StyleGRU after resizing it to be $\mathcal{R}^{256\times 256\times l}$, while the raw clip $I$ is an input of 3D ResNet-50 module with the resized shape of $\mathcal{R}^{224\times 224\times l}$.

\subsubsection{StyleGRU module}

The preprocessed video clip $I_p$ serves as the input to extract style latent vectors via the pSp encoder~\cite{pSp2021}.
The pSp encoder ($pSp$) provides the capability to extract style latent vectors for inverting given images within a well-trained StyleGAN~\cite{StyleGAN2} latent space.
We extract the style latent vectors $\mathbf{S}$ as:
\begin{equation}\label{S_tk}
    \mathbf{S} = pSp(I_{p}) \equiv \{s_{1}, s_{2}, ..., s_{l}\},
\end{equation}
where $s_{k} \in \mathcal{R^{D}}$ is the $D$-dimensional style vector of the $k$-th frame of the clip $I_p$, $k \in [1, l]$, and $l$ is the number of frames in clip. 

The style latent vector that we extract through the pSp encoder is a feature for the GAN inversion task, so it is insufficient for direct application to the deepfake detection task.
To enable the extension of style from the image domain to the video, we employ the difference of subsequent style latent vectors, extracted from frames within a clip.
The style flow $\Delta s$, derived from difference of style latent vectors, is represented as $\Delta s_{i} = s_{i+1} - s_{i}$.
Then, the input $\Delta \mathbf{S}$, which is called \textit{style flow}, is constructed as follows:
\begin{equation}\label{F_tk}
    \Delta \mathbf{S} = \{\Delta s_{1}, \Delta s_{2}, ..., \Delta s_{(l-1)}\},
\end{equation}

The style flow $\Delta \mathbf{S}$ is an input to the Gated Recurrent Unit (GRU) layers~\cite{gru}.
Then, we obtain a temporal embedded style flow $E_{\text{style}}\in \mathcal{R}^{C\times B\times L}$ through the hidden state of the final GRU layer.
$C$ represents the hidden unit size of GRU, $B$ indicates the number of directions of GRU, and $L$ denotes the number of GRU layers.
We build two consecutive GRU layers of 4096 hidden units with bidirectional configuration, so $C=4096$, $B=2$, and $L=2$.
Thus, we can derive that:
\begin{equation}\label{E}
    E_{\text{style}} = \mathbf{GRU}(\Delta \mathbf{S}),
\end{equation}
where $\mathbf{GRU}$ represents the operation of bidirectional two GRU layers embedded in StyleGRU module.

\subsubsection{3D ResNet-50 module}

The content feature $C_{\text{content}}$ is derived from the feature extraction performed by the 3D CNN as:
\begin{equation}\label{C}
    C_{\text{content}} = \mathbf{3D~CNN}(I),
\end{equation}
where $\mathbf{3D~CNN}$ is an operation function for extracting a feature from 3D CNN model.
Thus, $C_{\text{content}}$ contains information about visual and temporal inconsistency.
We utilize the feature of $\mathcal{R}^{1024\times 16}$ by applying a global average pooling to the last convolution layer of 3D-R50 architecture~\cite{3DR50-2017} pre-trained in FTCN~\cite{FTCN2021}.

\subsubsection{Style attention module}

We introduce a Style Attention Module (SAM) that leverages the style embedding vector obtained from the StyleGRU.
To apply the attention mechanism, we designate $C_{\text{content}}$ as the query, while $E_{\text{style}}$ is allocated as the key and value.
We employ the content feature as the query based on the intuition that these features hold the potential to identify anomalies in the style flow, driven by their encoded temporal inconsistency.

Thus, we design the query, key, and value of SAM as:
\begin{equation}\label{qkv}
Q_{C} = \mathbf{\phi}_{q}(C_\text{content}),~ K_{E} = \mathbf{\phi}_{k}(E_\text{style}),~ V_{E} = \mathbf{\phi}_{v}(E_\text{style}),
\end{equation}
where $\phi_q$, $\phi_k$, and $\phi_v$ are linear projection function using fully connected layer to estimate the query, key, and value, respectively.
Then, we calculate a Style Attention (SA) using the query and key as:
\begin{equation}\label{SA}
    \mathbf{SA}(Q_{C}, K_{E}) = s\left(\frac{Q_{C}K_{E}}{\sqrt{d_{K_{E}}}}\right),
\end{equation}
where $s(\bullet)$ represents a softmax function.

Ultimately, we compute the result of SAM by taking the dot-product of the SA with the value component.
Thus,
\begin{equation}\label{SAM}
\mathbf{SAM}(E, C) = \phi\big(\mathbf{SA}(Q_{C}, K_{E}) \cdot V_{E}\big),
\end{equation}
where $\phi$ is the last linear projection layer of SAM.

\subsubsection{Temporal transformer encoder module.} 
The output from the SAM is linearly projected into 16 time steps (n=16) and then fed into the Temporal Transformer Encoder (TTE) module.
The core of TTE is composed of standard transformer encoder blocks~\cite{transformer2017}, with each block including a multi-head self-attention module and multi-layer perceptrons.
The results of TTE are the input to the following classification head, yielding the final deepfake prediction through a sigmoid function $\sigma$. Thus,
\begin{equation}\label{prediction}
\hat{y} = \sigma\big(f_{cls}(\mathbf{TTE}(C_{\text{content}} + \mathbf{SAM}(E_{\text{style}}, C_{\text{content}})))\big),
\end{equation}
where $f_{cls}$ is a function of fully-connected layer to represent the classification head.
We refer to FTCN~\cite{FTCN2021} for the architecture of TTE.

\subsection{Training Procedure}
The training procedure consists of two following stages. 
The first stage involves training a Gated Recurrent Unit (GRU) layers to enable encoding of the temporal style latent flow suitable for deepfake detection. 
In the following stage, we train the 3D CNN model for content embedding, the style attention module, and the subsequent modules for the binary classification.
Fig.~\ref{fig3} provides a comprehensive illustration of our training procedures.
\vspace{-0.3cm}

\subsubsection{Stage 1: style representation learning}
In stage 1, we train StyleGRU to learn the representation of style flow.
We employ supervised contrastive learning~\cite{supervised2020} to produce an effective representation of style flow.
For the supervised contrastive learning, we organized the style flows used as inputs to the GRU layers into three categories: anchor, positive, and negative style flows.

We utilize a different sampling strategy to build the clip for anchor, positive, and negative style flows.
At first, the anchor style flow ($\Delta\mathbf{S}_{a}$) is extracted from the clip sampled from the video sequence by a sliding window sampling strategy.
On the other hand, the positive style flow ($\Delta\mathbf{S}_{p}$) is obtained by using a random sampling strategy.
Both the anchor and the positive style flows are extracted from the preprocessed clips with the same label.
In contrast, the negative style flow ($\Delta\mathbf{S}_{n}$) is obtained from the clips with labels distinct from those of the anchor and positive clips.
The negative style flow is generated by randomly employing one of the two sampling strategies.

As a batch, the anchor, positive, and negative style flows are fed into the GRU layers, and their representation is trained by using a triplet loss and a classification loss.
When we define the GRU's final hidden states from $\Delta \mathbf{S}_{a}$, $\Delta \mathbf{S}_{p}$, and $\Delta \mathbf{S}_{n}$ by $E_{a}$, $E_{p}$, and $E_{n}$, respectively, the triplet loss for training the GRU layers is derived as follows:
\begin{equation}\label{L_tri}
    L_{tri} = \mathbf{max}\big(||E_{a} - E_{p}||_{2}^{2}- ||E_{a} - E_{n}||_{2}^{2} + \alpha, 0\big),
\end{equation}
where $\alpha$ is a margin parameter.

Additionally, we used classification loss ($L_{cls}$), which is designed by Binary Cross-Entropy (BCE) loss to learn appropriate features for classification.
To obtain the predicted value $\hat{y}$ for the ground truth label $y$, the hidden states $E$ of the GRU are input into the auxiliary classifier.
The calculated BCE($y$, $\hat{y}$) loss is utilized as $L_{cls}$.
The auxiliary classifier consists of three fully connected layers, with a residual connection applied after the first two layers.
The hidden size of each fully connected layer is kept identical at 4096.

Then, the entire loss function for training StyleGRU is as follows:
\begin{equation}\label{L_total}
    L = L_{tri} + \lambda L_{cls},
\end{equation}
where $\lambda$ is a user-defined hyperparameter to determine the influence degree of $L_{cls}$.
We minimize the loss by using Adam optimizer~\cite{adam2014}.

\subsubsection{Stage 2: style attention-based deepfake detection.}
In stage 2, to perform the real/fake binary classification task, we employ a conventional Binary Cross-Entropy (BCE) loss to train the 3D ResNet-50, SAM, and TTE modules, while fixing the StyleGRU module trained at stage 1.
The BCE loss ($L_{bce}$) is applied to reduce the gap between the fake prediction $\hat{y}$ of Eq.~\ref{prediction} and its corresponding ground truth label, where 0 is real and 1 is fake.
We used the momentum-driven SGD optimizer for the training process of stage 2.

\section{Experiments}

We conduct experiments in this section to demonstrate the superiority of our algorithm.
Experiments under cross-dataset and cross-manipulation conditions serve as benchmark settings to assess the generalization performance of deepfake detection models.

\noindent\textbf{Implementation Details. }
During the preprocessing phase, we utilized Dlib~\cite{dlib2009} to extract facial landmarks and perform alignment, as well as Retinaface~\cite{retinaface2020} to carry out face cropping.
Following this preprocessing, the real and fake clips fed into the model consist of a total of $l=32$ frames each.
In stage 1, a dropout layer with a dropout rate of $p=0.2$ is included between the input style flow and the StyleGRU.
Furthermore, in StyleGRU, an RNN dropout rate of $p= 0.1$ is applied.
As training parameters, a total of 50,000 samples are organized with a batch size of 256 over 100 epochs for contrastive learning.
For optimization, the ADAM optimizer is used with a learning rate of $5e^{-4}$.

In stage 2, we use the momentum SGD optimizer and set the weight decay to $1e^{-4}$.
The training is conducted with a batch size of $16$.
For the first 10 epochs, the learning rate of SGD gradually warms up from $0.01$ to $0.1$.
Subsequently, a cosine learning rate scheduler is used to decrease the learning rate throughout 100 epochs.
\begin{table}[ht]
\vspace{0.2cm}
\centering
\resizebox{\columnwidth}{!}
{
\begin{tabular}{l | c c c c | c}
    \hline
    Method & CDF & DFD & FSh & DFo & Avg \\ 
    \hline
    Xception~\cite{xception2017}                    & 73.7      & -     & 72.0      & 84.5      & - \\
    CNN-aug~\cite{cnn-aug2020}                      & 75.6      & -     & 65.7      & 74.7      & - \\
    PatchForensics~\cite{patchforensics2020}        & 69.6      & -     & 57.8      & 81.8      & - \\
    Multi-task~\cite{multitask2019}                 & 75.7      & -     & 66.0      & 77.7      & - \\
    FWA~\cite{fwa2018}                              & 69.5      & -     & 65.5      & 50.2      & - \\
    Two-branch~\cite{twobranch2020}                 & 76.7      & -     & -         & -         & - \\
    DCL~\cite{dcl2022}                              & 82.3      & 91.6  & -         & -         & - \\
    Face X-ray~\cite{facexray2020}                  & 79.5      & \underline{95.4}  & 92.8      & 86.8      & 88.6 \\
    NoiseDF~\cite{noisedf2023}                      & 75.9      & -     & -         & 70.9      & - \\
    SLADD~\cite{sladd2022}                          & 79.7      & -     & -         & -         & - \\ 
    SBI-R50~\cite{sbi2022}                          & 85.7      & 94.0  & 78.2      & 91.4      & 87.3\\
    \hline
    CNN-GRU~\cite{cnn-gru2019}                      & 69.8      & -     & 80.8      & 74.1      & - \\
    STIL~\cite{stil2021}                            & 75.6      & -     & -         & -         & - \\
    LipForensics~\cite{lipforensics2021}            & 82.4      & -     & 97.1      & 97.6      & - \\
    RealForensics*~\cite{realforensics2022}         & 85.6      & 82.2  & \underline{\textbf{99.6}} & \underline{\textbf{99.8}} & 91.8 \\
    FTCN*~\cite{FTCN2021}                           & \underline{86.9}  & 94.4      & 98.8      & 98.8      & 94.7 \\
    AltFreeze*~\cite{altfreezing2023}               & \underline{\textbf{89.0}}     & 93.7      & \underline{99.2}      & \underline{99.0}      & \underline{95.2} \\
    \hline
    StyleGRU                                        & 62.5      & 72.2  & 90.7      & 77.7      & 79.8 \\
    Ours                                            & \underline{\textbf{89.0}}     & \underline{\textbf{96.1}}  & 99.0  & \underline{99.0}  & \underline{\textbf{95.8}} \\
    \hline
\end{tabular}
}
\caption{\textbf{Generalization to cross-dataset.} We present the AUC scores (\%) on the CDF, DFD, FSh, and DFo datasets to assess the cross-dataset performance of our framework. The highest-performing score is marked in \textbf{bold}. The \underline{underlined} values correspond to the Top-2 best methods. The asterisk (*) denotes that we have reproduced the results using officially provided weights.The results of other methods were obtained from AltFreeze~\cite{altfreezing2023}.
}
\label{table1}
\vspace{-0.4cm}
\end{table}

\noindent\textbf{Datasets and Measurement.} To ensure robustness under various conditions, we used the following deepfake detection datasets:
(1) \textbf{FaceForensics++} (FF++) \cite{ff++2019} comprises 1,000 original videos and 4,000 fake videos produced in four manipulation methods, Deepfake (DF) \cite{Deepfakes}, Face2Face (F2F) \cite{Face2Face}, FaceSwap (FS) \cite{FaceSwap}, NeuralTexture (NT) \cite{Neuraltexture}, and we utilize the c23 version.
(2) \textbf{FaceShfiter} (FSh) \cite{faceshifter2019} and (3) \textbf{DeeperForensics} \cite{DFo} result from the employment of optimized manipulation techniques to the original videos of FF++.
We executed our experiments using test videos adhering to the same split as in FF++.
(4) \textbf{CelebDF-v2} (CDF) \cite{CDF} is composed of 590 real videos and 5,639 fake videos.
For our study, we engaged a subset of 518 test videos.
(5) \textbf{DeepfakeDetection} (DFD) \cite{DFD} consists of 363 real videos and 3071 synthetic videos.

To train our model, we used the training videos from the FF++~\cite{ff++2019} dataset with light compression, specifically identified as c23.
The evaluation metric is the Area Under the receiver operating characteristic Curve (AUC).
Most deepfake detection algorithms have utilized AUC as their evaluation metric, and Lipforensics \cite{lipforensics2021}, which is a fundamental benchmark for video-based deepfake detection, also employed AUC.

\noindent\textbf{Comparisons.} We have compared both image-based and video-based deepfake detection models.
Xception \cite{ff++2019} is a method of training the Xception model \cite{xception2017}, which is a widely used CNN architecture.
CNN-aug \cite{cnn-aug2020} discovered that CNN-generated images can be readily identified by a CNN model.
PatchForensics \cite{patchforensics2020} indicated that the use of patch-based classifiers can contribute to robust deepfake detection.
Multi-task \cite{multitask2019} approach employed an architecture similar to that of an auto-encoder for deepfake detection.
FWA \cite{fwa2018} introduced a training data generation technique that relies on deteriorating the quality of the GAN-generated source images.
Two-branch \cite{twobranch2020} method utilized multi-task learning within the context of a deepfake detection dataset.
SLADD \cite{sladd2022} introduced an algorithm that automatically learns effective augmentation techniques for deepfake detection.
SBI \cite{sbi2022} proposed a novel approach to blending artifacts using self-images.
DCL \cite{dcl2022} applied contrastive learning to the task of deepfake detection in the image domain.
NoiseDF \cite{noisedf2023} compared the features of the cropped face and background squares using noise to detect deepfakes.
Face X-ray \cite{facexray2020} identified face forgery by segmenting and delineating the blending border between source and target images.
CNN-GRU \cite{cnn-gru2019} highlighted the potential of utilizing CNN and GRU to investigate temporal features.
STIL \cite{stil2021} specifically focused on learning spatio-temporal inconsistency.
LipForensics \cite{lipforensics2021} exhibited robust generalization performance through the use of a trained model to capture the representation of natural mouth movement.
RealForensics \cite{realforensics2022} improved the ability to generalize using multimodal architecture.
FTCN \cite{FTCN2021} argued that leveraging temporal inconsistency is beneficial for enhancing the generalizability of deepfake detection algorithms.
AltFreeze~\cite{altfreezing2023} is an architecture that demonstrates good generalization performance by separately training spatial information and temporal information.

\subsection{Generalization to cross-datasets}

In the context of our framework, we have set the cross-dataset performance as a proxy for the generalizability in real-world situations.
We train the proposed framework on the FF++ and then evaluate its performance on the CDF, DFD, FSh, and DFo datasets.
Table~\ref{table1} shows that our framework exhibits the best detection performance in the CDF, DFD, and DFo datasets, which include high-quality deepfake videos.
The performance on the FSh dataset also reaches a top-2 level, with the highest average score, indicating that our model is a deepfake detection algorithm with generalization capability.
`StyleGRU' represents the cross-dataset score using the classification layer directly connected after the StyleGRU for the Stage 1 training.
The score of StyleGRU is superior to that of CNN-GRU~\cite{cnn-gru2019}, which is a similar structure that utilizes image features, on the FSh and DFo datasets.
This suggests the potential for a better impact on generalization performance by utilizing the flow of style latent vectors rather than simply relying on the image-based feature.
\begin{table}[t]
\centering
\resizebox{\columnwidth}{!}
{
\begin{tabular}{l|cccc|c}
    \hline
    \multicolumn{1}{c|}{} & \multicolumn{4}{|c}{Train on remaining three}\\
    \multicolumn{1}{c|}{Method} & DF & FS & F2F & NT & Avg \\
    \hline
    Xception~\cite{xception2017}  & 93.9 & 51.2 & 86.8 & 79.7 & 77.9 \\
    CNN-aug~\cite{cnn-aug2020} & 87.5 & 56.3 & 80.1 & 67.8 & 72.9 \\
    PatchForensics~\cite{patchforensics2020} & 94.0 & 60.5 & 87.3 & 84.8 & 81.6 \\
    Face X-ray~\cite{facexray2020} & \underline{99.5} & 93.2 & 94.5 & 92.5 & 94.9 \\
    \hline
    CNN-GRU~\cite{cnn-gru2019} & 97.6 & 47.6 & 85.8 & 86.6 & 79.4 \\
    LipForensics~\cite{lipforensics2021} & \underline{99.7} & 90.1 & \underline{98.8} & \underline{98.3} & 97.1 \\
    FTCN*~\cite{FTCN2021}& \underline{99.8} & \underline{99.3} & 95.9 & 95.3 & \underline{97.5} \\
    AltFreeze~\cite{altfreezing2023} & \underline{99.8} & \underline{99.7} & \underline{98.6} & \underline{96.2} & \underline{98.6} \\
    \hline
    StyleGRU & 94.0 & 68.5 & 88.8 & 80.6 & 83.0 \\
    Ours & \underline{99.7} & \underline{98.8} & \underline{98.6} & \underline{96.4} & \underline{98.4} \\
    \hline
\end{tabular}
}
\caption{\textbf{Generalization to cross-manipulation.} In order to quantitatively assess the performance of our method in cross-manipulation scenarios, we utilized video-level AUC. The asterisk (*) denotes results that we have reproduced. The \underline{underlined} values correspond to the Top-3 best methods. The results of other methods were obtained from AltFreeze~\cite{altfreezing2023}.}
\label{table2}
\end{table}

\subsection{Generalization to cross-manipulations}
For deepfake detection tasks, the evaluation of cross-manipulation performance targets fake videos generated from the same original video but using a different scheme.
A common approach is to select three out of the four manipulations provided by FF++, and then the model trained by the three chosen manipulation sets is evaluated by one remaining manipulation set.
DF~\cite{Deepfakes} and FS~\cite{FaceSwap} represent identity swap manipulation methods, while F2F~\cite{Face2Face} and NT~\cite{Neuraltexture} denote expression swap manipulation methods.
AltFreeze~\cite{altfreezing2023} does not officially provide weights for this experiment setting.
Therefore, it is not possible for us to reproduce.
Table~\ref{table2} presents a comparison of the performance of the cross-manipulation between our model and various models.
For expression swap manipulation methods like F2F and NT, they exhibit relatively lower performance compared to Lipforensics~\cite{lipforensics2021}.
Expression swap inherently lead to inconsistencies in the mouth region, which is why LipForensics appears to perform well.
However, our framework displays outstanding performance across all manipulations.

\subsection{Ablation Study}
In this section, we conduct ablation studies to verify the approach used in our framework.
we validate the necessity of the proposed StyleGRU and SAM and conduct an ablation study on the losses used in Stage 1.
\vspace{-1mm}
\subsubsection{Component-wise Ablation Study}
As shown in Table~\ref{table3}, we conducted experiments using only the StyleGRU by feeding the hidden states of StyleGRU into an MLP classification head to detect the real/fake of input video.
Even though only the classification through StyleGRU was conducted, ours still demonstrates high scores on the Fsh dataset.
In addition, considering the significant performance drop when StyleGRU is not used, it can be observed that our proposed StyleGRU contributes to improving cross-dataset performance.
The influence of the proposed SAM in Stage 2 on the outcome is also evaluated.
For experiments without StyleGRU, we processed style latent vectors extracted from clips using an MLP to match the dimension of the 3D CNN features.
In the case of experiments without the SAM, we simply added the StyleGRU feature to the 3D CNN feature.
The absence of the SAM during the blending of content and style features leads to a substantial decline in performance.
This quantitatively demonstrates the efficacy of the SAM trained during stage 2 in achieving the adapted feature for robust deepfake detection.
Finally, the influence of differentiating ($\Delta$) the style latent vectors is investigated by replacing the input of the style flow with the original style latent vectors.
During the overall stage 1 training process, it was observed that the lack of differencing in style latent vectors as a pre-processing to be fed into the StyleGRU leads to a slower decrease in the loss.
Training without differentiation was confirmed to not only lead to slower progression in stage 1 learning but also adversely affect cross-dataset performance.
\vspace{-0.28cm}
\begin{table}[t]
\centering
\begin{tabular}{cccc|ccc}
\hline
\multicolumn{4}{c|}{Method}  & \multicolumn{2}{c}{Dataset}            \\
StyleGRU & SAM & TTE & $\Delta$ & CDF  & \multicolumn{1}{c|}{FSh}  & Avg \\ \hline
\checkmark        & -   & -   & \checkmark     & 62.5 & \multicolumn{1}{c|}{90.7} & 76.6    \\
-        & \checkmark   & \checkmark   & \checkmark     & 85.7 & \multicolumn{1}{c|}{98.4} &  92.1   \\
\checkmark        & -   & \checkmark   & \checkmark     & 87.8 & \multicolumn{1}{c|}{98.3} &  93.1   \\
\checkmark        & \checkmark   & \checkmark   & -     & 88.4 & \multicolumn{1}{c|}{98.4} &  93.4   \\
\checkmark        & \checkmark   & \checkmark   & \checkmark     & 89.0 & \multicolumn{1}{c|}{99.0} &  94.0    \\ \hline
\end{tabular}
\caption{\textbf{Framework ablation.} We evaluate our framework on the CDF and FSh datasets using video-level AUC scores and conduct ablation studies for each stage of the proposed methodology. All models utilized the FF++ dataset for training.}
\label{table3}
\end{table}

\begin{table}[t]
\vspace{-1mm}
\centering
\begin{tabular}{cc|ccc}
\hline
\multicolumn{2}{c|}{Loss} & \multicolumn{3}{c}{Dataset}             \\ \cline{3-5} 
$L_{cls}$      & $L_{tri}$   & \multicolumn{1}{c|}{FF++} & CDF  & FSh  \\ \hline
\checkmark     & -           & \multicolumn{1}{c|}{99.2} & 88.6 & 99.0 \\
-              & \checkmark  & \multicolumn{1}{c|}{98.8} & 88.4 & 98.8 \\
\checkmark     & \checkmark  & \multicolumn{1}{c|}{99.1} & 89.0 & 99.0 \\ \hline
\end{tabular}
\caption{\textbf{Stage 1 loss ablation.} We conduct an ablation study for the losses employed in stage 1. Each experiment uses the different combinations of training losses for StyleGRU, while preserving the training process of stage 2. The model is trained by the FF++ dataset, and evaluated by video-level AUC.}
\label{table4}
\end{table}
\subsubsection{Ablation Study for Losses}
Ablation studies are conducted on various options of training losses that can be employed at stage 1 training for StyleGRU representation learning.
As shown in Table~\ref{table4}, when only classification loss ($L_{cls}$) is used, our model exhibits best performance on the training dataset, while suffering from the reduction of domain generality.
Using only triplet loss ($L_{tri}$) results in relatively reduced performance due to the lack of distinctiveness between the real and fake videos for classification.
Although employing both $L_{cls}$ and $L_{tri}$ results in a slight performance decrease on the seen dataset, we can see that their concurrent employment proves to be beneficial on the unseen datasets.
\begin{figure}[t]
\centering
\includegraphics[width=1.0\columnwidth]{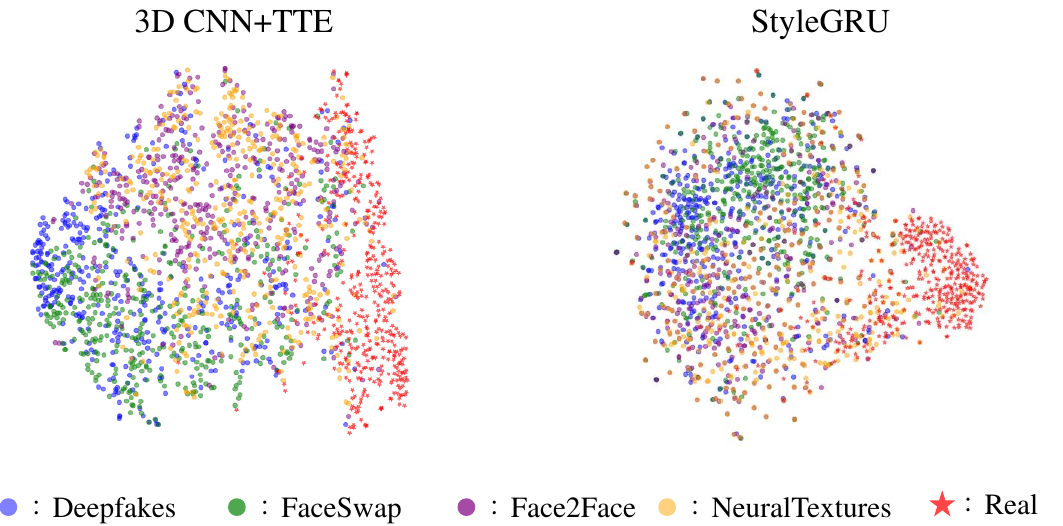} 
\caption{\textbf{t-SNE visualization.} The t-SNE visualization was conducted using the final layer features of a model trained on the FF++ test set. For hyperparameters, we utilized 1000 iterations with a perplexity of 40 and PCA to 30.}
\label{fig4}
\end{figure}

\subsection{Analytic Experiments}
In this section, we perform visualizations for further analysis of the proposed style latent flow in our study.
We also analyze the perturbation robustness of our algorithm, which is important to consider in real-world scenarios.
\vspace{-0.3cm}
\subsubsection{t-SNE for Domain Generalization}
To evaluate the learning performance of StyleGRU, we conduct t-SNE visualizations on the FF++ test set using a model trained on the FF++ dataset.
As illustrated in Fig.~\ref{fig4}, despite StyleGRU utilizing significantly fewer parameters, our model manages to extract more distinct features in the feature space.
This characteristic represents our intuition that the distinctiveness of style latent vectors aligns well with the generality of deepfake video detection tasks across various generation algorithms.

\subsubsection{Robustness to Unseen Perturbation}
\begin{figure}[h]
\centering
\includegraphics[width=1.0\columnwidth]{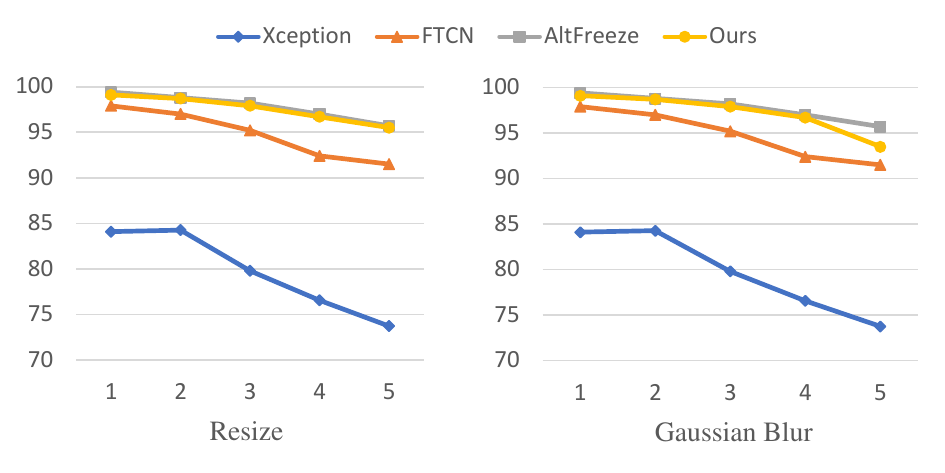} 
\caption{\textbf{Robustness to Unseen Perturbation.} The performance changes based on the video-level AUC metric when applying two perturbations at five different degradation levels. The perturbation follows the approach provided by DeeperForensics~\cite{DFo}.}
\label{fig5}
\end{figure}

To evaluate the robustness of perturbation, experiments under the conditions proposed by DFo~\cite{DFo} are conducted to observe performance changes.
In Fig.~\ref{fig5}, performance under specific perturbation conditions exhibits a slight deficiency, which is attributed to the fact that the pSp encoder~\cite{pSp2021} used for extracting style latent vectors was created without considering noise.
However, our model demonstrates a high level of robustness against most perturbations.

\section{Conclusion}
In this paper, we first demonstrate that the style latent flow extracted from consecutive frames of a video, serves as a effective cue for deepfake detection.
Additionally, we confirm that leveraging an attention mechanism based on style flow facilitates model generalization.
The style flow feature is trained via a supervised-contrastive representation learning process, and we develop a classification model using the Style Attention Module.
The superior performance of the proposed algorithm validates the importance of temporal changes in style latent vectors to generalize the deepfake video detection task.
While this paper focused on the facial attributes extracted by StyleGAN, our future plans is to broaden the scope of attributes for deepfake video detection to encompass various objects.
We anticipate that utilizing StyleGAN pre-trained on other subjects, such as images of dogs and scenery, could broaden the applicability of our method to a wider range of deepfake subjects.

The model we propose includes an additional step of extracting style latent vectors through the pSp encoder.
As a result, the data preprocessing takes a relatively long time.
However, we confirms the positive impact of using style flow on the generalization performance of deepfake video detection models and provides a new direction for addressing synthesized videos created through GAN inversion.

\vspace{1mm}
\begin{spacing}{0.8}
\footnotesize
\noindent\textbf{Acknowledgements:}
This work was partly supported by Institute of Information \& communications Technology Planning \& Evaluation (IITP) grant funded by the Korea government(MSIT) (2021-0-01341, Artificial Intelligence Graduate School Program(Chung-Ang University), 2020-0-01336, AIGS program (UNIST),  and 2021-0-01778 Development of human image synthesis and discrimination technology below the perceptual threshold) and Culture, Sports and Tourism R\&D Program through the Korea Creative Content Agency grant funded by the Ministry of Culture, Sports and Tourism in 2024 (Project Name : Development of high-freedom large-scale user interaction technology using multiple projection spaces to overcome low-light lighting environments. , Project Number : RS-2023-00222280).
\normalsize
\end{spacing}

\clearpage
\setcounter{page}{1}
\maketitlesupplementary
The supplementary details provide specific explanations about the settings that were not fully described in the main paper, including various configuration values.
Additionally, we present supplementary information that covers experimental results and visualizations which could not be included in the paper due to space limitations.
As a result, we provide additional support to the validity of the results presented in our original paper.

\section{More Implementation Details}
\label{sec:More Implementation Details}
We employed two preprocessing methods to detect and crop faces in the videos. 
First, we use RetinaFace~\cite{retinaface2020} to detect and align the faces for each video. 
We use landmarks to determine the average face region and then use that region to crop the faces. 
Each clip consists of 32 cropped faces resized to 224$\times$ 224 and is used as input for the 3D CNN in Stage 2. 

All experiments were conducted using four Nvidia A6000 48GB GPUs and an AMD Ryzen Threadripper PRO 3955WX 16-Cores CPU.

\subsection{Stage 1}
In Stage 1, we perform preprocessing to prepare the input for the pre-trained pSp encoder.
We align and crop the faces using dlib\cite{dlib2009}.
These face images are resized to 256$\times$256.

We use the total loss $L$ in training StyleGRU, where $\lambda$ sets to 1.
\begin{equation}
\label{supp:L_total}
L = L_{tri} + \lambda L_{cls},
\end{equation}

\subsection{Stage 2}
We conduct data augmentation through the cutout. 
When applying the cutout, n square regions are randomly selected, ranging in size from 20\% to 80\% of the total image area.
These cutout regions are applied uniformly to all frames within the clip.
We employed the same 3D CNN architecture and Temporal Transformer Encoder structure as FTCN~\cite{FTCN2021}.
For cross-dataset experiments, we utilized the pretrain weights provided by the existing FTCN model to facilitate effective learning.
During this process, we employed the SGD optimizer with momentum and trained with a learning rate of 5e-7.

\section{Additional Experiments}
\subsection{Style Latent Selection}
\begin{table}[h]
\centering
\begin{tabular}{l|cc}
    \hline
    Latent & CDF & FSh \\
    \hline
     coarse & 88.1 & 98.5 \\
     middle & 87.8 & 98.5 \\
     fine   & 88.1 & 98.7 \\
     total (Ours)  & 89.0 & 99.0 \\
    \hline
\end{tabular}
\caption{\textbf{Style Latent ablation study.}}
\vspace{-0.5em}
\label{table1_sup}
\end{table}

According to \cite{latentforensics2023}, training with specific latents is effective for deepfake detection.
To test this argument for our model, we divide the extracted \textbf{total} style latent (18$\times$512) into \textbf{coarse} (3$\times$512), \textbf{middle} (4$\times$512), and \textbf{fine} (11$\times$512) segments, following the suggestion of  StyleGAN\cite{StyleGAN1}, and conduct separate experiments on each segment. 
To evaluate our model's generalization performance, we train it on the FF++~\cite{ff++2019} dataset and then conduct performance evaluations on the CDF~\cite{CDF} and FSh~\cite{FaceShifter} datasets.

Our model's ablation study performance undergoes assessment through the presentation of AUC scores ($\%$) on the coarse, middle, fine, and total style latent vectors. In Table~\ref{table1_sup}, we can observe that utilizing all latent variables ultimately demonstrates higher generalization performance.

\begin{table}[h]
\centering
\begin{tabular}{l|ll}
\hline
Metric                        & CDF  & Fsh  \\ \hline
No differencing               & 88.4 & 98.4 \\
2nd-order differencing        & 88.8 & 98.8 \\ \hline
Ours(1st-order differencing)  & \textbf{89.0} & \textbf{99.0} \\ \hline
\end{tabular}
\vspace{-0.3cm}
\caption{\textbf{Flow modeling metric comparison experiment.}}
\vspace{-0.5em}
\label{table2_sup}
\end{table}
As presented in Table~\ref{table2_sup}, it reveals that the StyleGRU achieves superior performance when utilizing a metric based on first-order differencing, compared to other approaches such as neglecting differencing or employing second-order differencing. Nonetheless, the small performance gaps suggest that the GRU layer lets the style feature independent of the metric.

\begin{table}[h]
\centering
\begin{tabular}{l|cc}
    \hline
    Condition & CDF & FSh \\
    \hline
     Self-supervised           & 88.5 & 98.4 \\
     Supervised (Ours)    & 89.0 & 99.0 \\
    \hline
\end{tabular}
\caption{\textbf{Contrastive learning ablation study.}}
\vspace{-0.5cm}
\label{table3_sup}
\end{table}
\begin{table*}[t]
\centering
\begin{tabular}{l|c|ccccccc|c}
\hline
Method       & \multicolumn{1}{c|}{Clean} & \multicolumn{1}{c}{Saturation} & \multicolumn{1}{c}{Contrast} & \multicolumn{1}{c}{Block} & \multicolumn{1}{c}{Noise} & \multicolumn{1}{c}{Blur} & \multicolumn{1}{c}{Pixel} & \multicolumn{1}{c|}{Compress} & \multicolumn{1}{c}{Avg} \\ \hline
Xception~\cite{xception2017}     & 99.8                       & 99.3                           & 98.6                         & 99.7                      & 53.8                      & 60.2                     & 74.2                      & 62.1                          & 78.3                    \\
CNN-agu~\cite{cnn-aug2020}      & 99.8                       & 99.3                           & 99.1                         & 95.2                      & 54.7                      & 76.5                     & 91.2                      & 72.5                          & 84.1                    \\
Patch-based~\cite{patchforensics2020}  & 99.9                       & 84.3                           & 74.2                         & 99.2                      & 50.0                      & 54.4                     & 56.7                      & 53.4                          & 67.5                    \\
CNN-GRU~\cite{cnn-gru2019}      & 99.9                       & 99.0                           & 98.8                         & 97.9                      & 47.9                      & 71.5                     & 86.5                      & 74.5                          & 82.3                    \\
FTCN*~\cite{FTCN2021}        & 99.5                       & 98.0                           & 93.7                         & 90.1                      & 53.8                      & 95.0                     & 94.8                      & 83.7                          & 87.0                    \\
AltFreeze*~\cite{altfreezing2023} & 99.8                       & 99.4                           & 98.9                         & 91.8                      & 60.9                      & 98.3                     & 97.8                      & 89.9                          & 91.0                    \\ \hline
Ours         & 99.6                       & 99.2                           & 95.8                         & 92.2                      & 55.0                        & 97.3                     & 97.3                      & 86.3                          & 90.4                    \\ \hline
\end{tabular}
\caption{\textbf{Robustness to Perturbations.} We evaluate the average performance change based on the video-level AUC scores when applying distortions at five different degradation levels. The asterisk(*) denotes that we have reproduced the results using officially provided weights. The perturbation follows the approach provided by DeeperForensics~\cite{DFo}.}
\label{table4_sup}
\vspace{-1.0em}
\end{table*}
\subsection{Contrastive learning comparison}
Training the StyleGRU in Stage 1 applies a supervised contrastive learning manner using anchor, positive and negative samples with given labels.
On the other hand, for representation learning, self-supervised contrasitive learning approaches without providing labels are often employed.
We compared the performance difference between our supervised contrastive learning and a self-supervised contrastive learning approach in which clips extracted from the same video as the anchor clip were used as positive clips, and clips from different videos were used as negative clips, without using label. We conduct an ablation study on the representation learning methodology. We train on the FF++ dataset and verify performance through video-level AUC scores on the CDF and FSh datasets.

Table~\ref{table3_sup} illustrates that when employing supervised contrastive learning, it exhibits superior generalization performance compared to models utilizing self-supervised contrastive learning, thereby showcasing the efficacy of our approach.

\subsection{More about Perturbation Robustness}
We conduct a comparison study of the robustness of our proposed model to perturbations, which was not adequately presented in the main paper due to space constraints.
As Table~\ref{table4_sup} illustrates, it demonstrates suboptimal performance for a specific perturbation.
Specifically, performance is vulnerable to ` Noise ' type perturbations, which is attributed to the relatively sensitive response of the pSp encoder~\cite{pSp2021} to noise.
However, it exhibits a high level of performance compared to the majority of methods, and there is potential for improvement through future model enhancements.

\begin{table}[h]
\centering
\begin{tabular}{c|ccc}
\hline
Method   & raw      & c23     & c40     \\ \hline
FTCN*     & 99.5     & 99.0     & 70.2    \\ \hline
Ours     & \textbf{99.6}     & \textbf{99.1}    & \textbf{71.8}    \\ \hline
\end{tabular}
\caption{\textbf{Experiment on low-quality images.} }
\vspace{-0.5em}
\label{table5_sup}
\end{table}
As shown in Table~\ref{table5_sup}, we performed an experiment to verify if our algorithm generalizes well to low-quality samples even when trained on high-quality ones.
In the results, it can be observed that our method exhibits relatively robust performance against low-quality samples.

\begin{table}[h]
\centering
\begin{tabular}{l|ccc|c}
\hline
Method    & CDF  & DFD  & KoDF      & Avg     \\ \hline
FTCN*      & 86.9 & 94.4 &  69.4      & 83.6    \\
Altfreeze* & \textbf{89.0} & 93.7 &  69.8      & 84.2    \\ \hline
Ours      & \textbf{89.0} & \textbf{96.1} &  \textbf{71.1}      & \textbf{85.4}    \\ \hline
\end{tabular}
\caption{\textbf{Generalization to distinct datasets from training data.}}
\label{table6_sup}
\end{table}
\begin{figure}[ht]
\centering
\includegraphics[width=1.0\columnwidth]{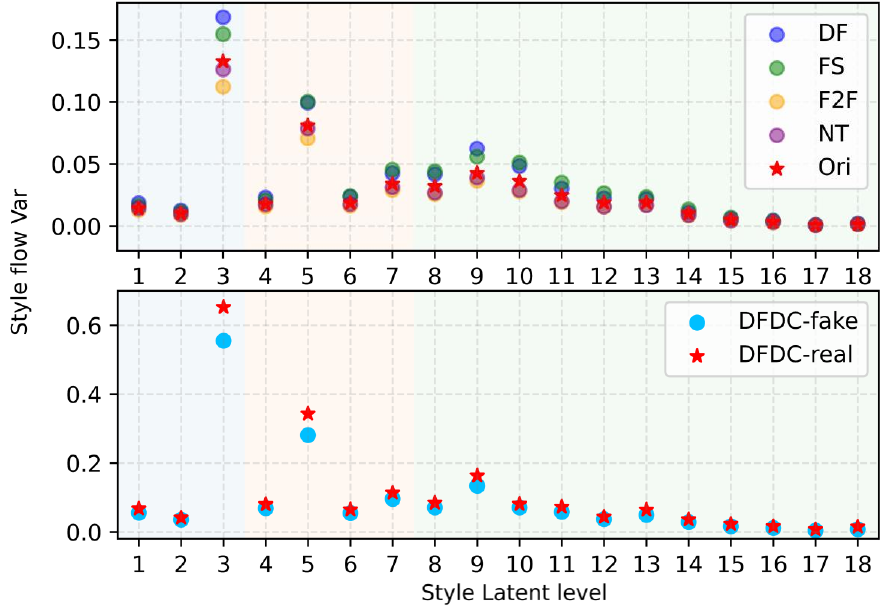} 
\caption{\textbf{Visualization about Style Latent variance.} The x-axis represents the levels of style latent vectors for detailed style representations.The plot above visualizes the results for the FF++ dataset, which we primarily used as the Train dataset, and the plot below presents the visualization results for the DFDC dataset.}
\label{fig1}
\end{figure}

\section{Visualization}
\subsection{More about style latent variance}
We provide additional explanations regarding the motivation behind the variance in style latent and conduct experiments on different datasets in our paper.
We utilized the style latent vectors with differencing from the first clip of each video in the dataset. To examine temporal changes, we computed the variance and visualized it as a means of statistical analysis.

According to the results in Figure~\ref{fig1}, a noticeable distinction between real and fake videos is evident in both the FF++~\cite{ff++2019} and DFDC\cite{dfdc} datasets.
The DF~\cite{Deepfakes}, FS~\cite{FaceSwap}, and DFDC datasets, which involve the identity swap method, exhibit differences in a manner similar to what was observed in the visualizations of DFD~\cite{DFD} and CDF~\cite{CDF}. On the other hand, the F2F~\cite{Face2Face} and NT~\cite{Neuraltexture} datasets, which involve the expression swap method, show reduced variance due to the fixed identity throughout the manipulation.

\subsection{Qualitative comparison with SAM score}
\begin{figure}[t]
\centering
\includegraphics[width=1.0\columnwidth]{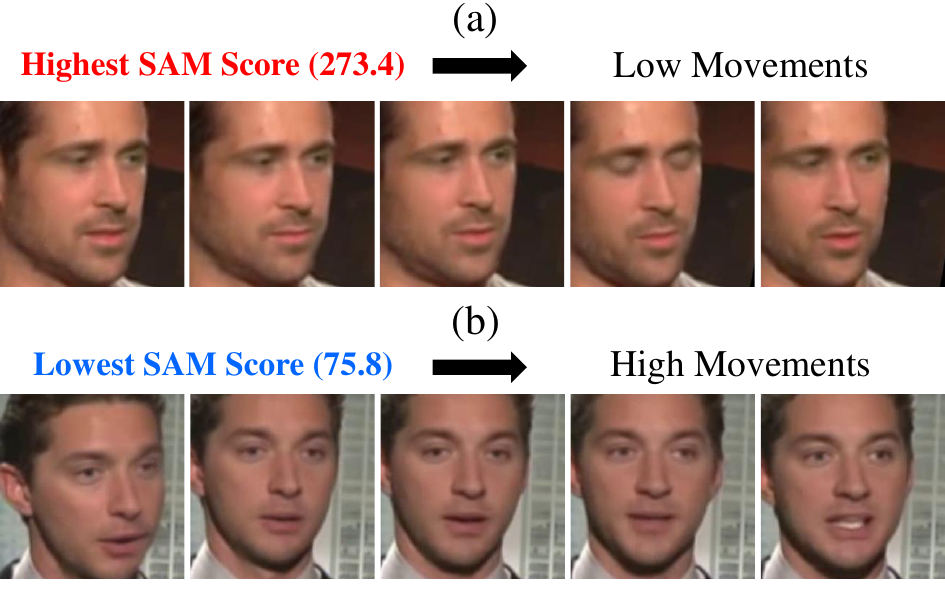} 
\caption{\textbf{Qualitative comparison with SAM score.} The displayed frames are extracted from correctly classified synthesis videos in the CDF dataset, with each frame taken at intervals of 6 frames within a single clip.}
\label{fig2_sup}
\end{figure}

In Figure~\ref{fig2_sup}, a qualitative evaluation is carried out for the SAM scores.
Figure~\ref{fig2_sup}(a) illustrates frames extracted from fake clips with notable SAM scores.
When considering classification based on low-level temporal cues, it becomes challenging to classify clips demonstrating slow movement.
The SAM proposed in our work generates high responses for clips containing slow movement, actively utilizing high-level temporal cues.
On the other hand, as shown in Figure~\ref{fig2_sup}(b), for clips with significant motion, it is observed that low responses are generated to focus on low-level temporal cues.
This result suggests that the style latent vector flow proposed in this study can be used as a complementary feature to the conventional image-based temporal artifact. 

{
    \small
    \bibliographystyle{ieeenat_fullname}
    \bibliography{main}
}

\end{document}